\documentclass[11pt]{article}

\usepackage[margin=1.25in]{geometry}

\usepackage[utf8]{inputenc} 
\usepackage[T1]{fontenc}    
\usepackage{microtype}      

\usepackage{graphicx}
\usepackage{subcaption}
\usepackage{booktabs}       

\usepackage{amsmath}
\usepackage{amssymb}
\usepackage{mathtools}
\usepackage{amsthm}

\usepackage[authoryear,round,sort]{natbib}
\usepackage{algorithm}
\usepackage{algorithmic}

\renewcommand{\cite}{\citep}

\usepackage{authblk}

\usepackage{xcolor}
\definecolor{darkblue}{rgb}{0.0, 0.08, 0.55}

\usepackage[
  colorlinks=true,
  linkcolor=darkblue,
  citecolor=darkblue,
  urlcolor=darkblue,
  pdfborder={0 0 0}
]{hyperref}

\usepackage[capitalize,noabbrev]{cleveref}

\theoremstyle{plain}
\newtheorem{theorem}{Theorem}[section]

\theoremstyle{definition}
\newtheorem{definition}[theorem]{Definition}

\theoremstyle{remark}

\usepackage[disable,textsize=tiny]{todonotes}

\title{STACHE: Local Black-Box Explanations for \\ Reinforcement Learning Policies}

\author[]{Andrew Elashkin}
\author[]{Orna Grumberg}

\affil[]{Faculty of Computer Science, Technion -- Israel Institute of Technology}

\date{}

\begin{document}

\maketitle

\begin{abstract}
Reinforcement learning agents often behave unexpectedly in sparse-reward or safety-critical environments, creating a strong need for reliable debugging and verification tools. In this paper, we propose STACHE, a comprehensive framework for generating local, black-box explanations for an agent's specific action within discrete Markov games. Our method produces a \textit{Composite Explanation} consisting of two complementary components: (1) a \textit{Robustness Region}, the connected neighborhood of states where the agent's action remains invariant, and (2) \textit{Minimal Counterfactuals}, the smallest state perturbations required to alter that decision. By exploiting the structure of factored state spaces, we introduce an exact, search-based algorithm that circumvents the fidelity gaps of surrogate models. Empirical validation on Gymnasium environments demonstrates that our framework not only explains policy actions, but also effectively captures the evolution of policy logic during training — from erratic, unstable behavior to optimized, robust strategies — providing actionable insights into agent sensitivity and decision boundaries.
\end{abstract}

\section{Introduction}
\label{sec:introduction}

Despite the impressive advances made by deep reinforcement learning (RL) agents, their decision-making processes remain opaque \cite{Cheng2025SurveyXDRL,Qing2022XRLsurvey}. This "black-box" nature poses serious concerns for settings where trust and reliability are critical. Deploying RL agents requires ensuring they make decisions for the right reasons, yet standard metrics like cumulative reward do not reveal the logic behind individual actions \cite{Milani2022SurveyXRL}.

While much research focuses on explaining global policy behavior or summarizing trajectories, there are critical scenarios where explaining a \textit{single} action is paramount. For instance, an agent that generally performs well might make a sudden, catastrophic error—such as a taxi agent turning into a wall. Understanding the precise cause of such a decision requires local explainability methods that can isolate the specific state factors responsible.

In this paper, we address this challenge by establishing a framework for \textbf{Composite Explanations}. We argue that to fully understand an action $a$ taken in state $s$, one must answer two questions: "How stable is this decision?" and "What would make it change?". To this end, we combine two analytical constructs:
\begin{itemize}
    \item \textbf{Robustness Regions:} The set of states in the local neighborhood of $s$ where the agent's policy remains invariant. This quantifies stability and reveals which factors the agent is ignoring (robustness) versus which it is strictly adhering to.
    \item \textbf{Minimal Counterfactuals:} The smallest perturbations to $s$ that trigger a change in action. This identifies the decision boundary and the specific features the agent is most sensitive to.
\end{itemize}

We make three main contributions: (1) We formalize a model-agnostic framework for local explanations in discrete Markov games, integrating \textit{Robustness Regions}—connected components of invariant behavior—with \textit{Minimal Counterfactuals} to simultaneously characterize decision stability and sensitivity without relying on policy approximations; (2) We introduce an exact, search-based algorithm that treats the policy purely as a black box—requiring no access to internal weights or gradients—to compute these explanations with 100\% fidelity to the agent's actual logic; and (3) We empirically demonstrate that our metrics effectively track the \textit{evolution of policy logic}, revealing that competent policies tend to develop narrow stability regions for actions requiring precision (like pickups) while growing broader, more stable regions for general navigation, offering a practical way to spot brittle behavior. 

\paragraph{Implementation:} The complete code for STACHE, including all experiments and visualization tools used in this paper, is available at \url{https://github.com/aelashkin/STACHE}.

\section{Related Work}
\label{sec:related-work}

Our work sits at the intersection of Explainable AI (XAI), Reinforcement Learning (RL), and robustness analysis. We distinguish our contribution by focusing on \textit{exact}, \textit{model-agnostic} explanations for discrete environments, contrasting with approximation-based or white-box approaches.

\paragraph{Explainable AI (XAI).} Early XAI focused on supervised learning. Feature attribution methods like LIME \cite{ribeiro2016why} and SHAP \cite{lundberg2017shap} approximate local behavior via surrogate models or Shapley values. While powerful, these methods provide scalar importance scores rather than concrete alternative states. Our work aligns with the "counterfactual" branch of XAI \cite{wachter2017counterfactual}, which offers contrastive explanations ("Why P rather than Q?"), arguing these are more cognitively accessible to humans \cite{miller2019explanation}.

\paragraph{Explainable RL (XRL).} Global XRL methods often distill policies into decision trees \cite{bastani2018viper} or programmatic policies \cite{verma2018pirl}. While interpretable, surrogates suffer from fidelity gaps. Local XRL methods include saliency maps \cite{greydanus2018visualizing}, which highlight visual attention but can be unreliable \cite{atrey2020exploratory}. Recent work has adapted counterfactuals to RL. \citet{olson2021counterfactual} generate counterfactual states for Atari, while \citet{amitai2023explaining} use visual outcome comparisons. Unlike methods requiring generative models (GANs) or causal graphs \cite{madumal2020causal}, our approach relies on exact search in factored state spaces, ensuring 100\% fidelity to the policy being explained.

\paragraph{Robustness in RL.} Traditional "Robust RL" aims to \textit{train} agents resilient to uncertainty \cite{Pinto2017RARL}. Our work differs fundamentally: we use robustness as an \textit{explanatory tool} to analyze a fixed policy post-hoc. We define robustness regions as connected components of action invariance, distinct from formal verification methods like adversarial perturbation bounds \cite{zhang2020robust} which often require white-box access.

\section{Preliminaries}
\label{sec:preliminaries}

We consider a reinforcement learning (RL) agent interacting with an environment modeled as a Markov game (or MDP in the single-agent case), defined by the tuple $\langle S, A, P, R, \gamma\rangle$. We assume the agent operates according to a deterministic policy $\pi: S \to A$.

\begin{definition}[Markov Game]
\todo[color=blue!20]{check if should be removed}
A Markov game is a tuple $\langle S, A, P, R, \gamma \rangle$, where $S$ is the state space, $A$ is the action space, $P$ is the transition function, $R$ is the reward function, and $\gamma$ is the discount factor.
\end{definition}

We focus on \textbf{factored state representations}, where each state $s \in S$ is a vector of $k$ distinct variables $(x_1, \dots, x_k)$. Each factor $X_j$ draws values from a domain $\mathcal{X}_j$. We classify factors into two types:
\begin{itemize}
    \item \textbf{Numerical:} $\mathcal{X}_j$ has a meaningful total order (e.g., coordinates, counts).
    \item \textbf{Categorical:} $\mathcal{X}_j$ has no intrinsic order; only equality is meaningful (e.g., colors, object types).
\end{itemize}

\subsection{User-Defined Factorization of a Markov Game}
\label{sec:user-defined-factorization}

Even when a Markov game environment admits a factored state representation, there is often more than one sensible way to choose the factors. Each way of partitioning the raw environment state into a set of typed variables ---what we call a \emph{user-defined factorization}--- determines how we define state similarity and what counts as an elementary change. This, in turn, shapes the state-similarity graph on which our explanation methods operate.

\begin{definition}[User-Defined Factorization \(\mathcal{F}\)]
\label{def:user-defined-factorization}
A \textbf{user-defined factorization \(\mathcal{F}\)} of a state space \(S\) is a mapping from each raw state \(s \in S\) to a tuple of \(k\) factor values \((x_1, x_2, \dots, x_k)\), comprised of:
\begin{enumerate}
    \item A set of \(k\) state variables (or factors) \(X_1, X_2, \dots, X_k\).
    \item For each factor \(X_j\), its corresponding domain \(\mathcal{X}_j\). The overall state is the Cartesian product \(\mathcal{X}_1 \times \mathcal{X}_2 \times \dots \times \mathcal{X}_k\).
    \item For each factor \(X_j\), an assigned type (Numerical or Categorical).
\end{enumerate}
The factorization \(\mathcal{F}\) specifies how each state is decomposed and interpreted for explanation. Different choices of \(\mathcal{F}\) can yield different distances between the same pair of raw states, and thus different sets of neighbors.
\end{definition}

\subsection{Hybrid Factored State Distance}
To define "minimal" changes and "neighborhoods," we require a distance metric $d: S \times S \to \mathbb{R}_{\ge 0}$. Standard metrics like Euclidean or Hamming distance are insufficient for mixed attribute spaces. We propose a hybrid metric:

\begin{definition}[Hybrid Factored State Distance]
\label{def:hybrid-distance}
Let $I_N$ and $I_C$ be the sets of indices for numerical and categorical factors, respectively. The distance between states $s=(x_1, \dots, x_k)$ and $s'=(x'_1, \dots, x'_k)$ is:
\[
d_{\text{hybrid}}(s, s') = \sum_{j \in I_N} |x_j - x'_j| + \sum_{j \in I_C} \mathbb{I}(x_j \neq x'_j)
\]
\end{definition}
This metric sums the Manhattan distance for numerical factors and the Hamming distance for categorical ones.

\textbf{Induced Graph Structure.}
The metric naturally induces a graph structure over the state space. Let $\delta_{\text{unit}}$ be the smallest positive distance definable by $d$ (typically 1 for discrete domains). We define the set of \textbf{immediate neighbors} of a state $s$ as $\mathcal{N}(s) = \{s' \in S \mid d(s, s') = \delta_{\text{unit}}\}$. The state space can thus be viewed as a graph $G=(S, E_d)$ where an edge $(s, s') \in E_d$ exists if $s' \in \mathcal{N}(s)$. Our explanation methods operate by exploring this graph.

\section{Methodology}
\label{sec:methodology}

We aim to explain a decision $\pi(s_0) = a_0$ made by a deterministic policy $\pi$. We rely only on query access to $\pi$.

\subsection{Robustness Regions}
The robustness region is defined not just by the distance, but rather by connectivity within the induced graph. We require two conditions: Action Invariance and Connectivity.

\begin{definition}[Robustness Region]
\label{def:robustness-region}
The robustness region $\mathcal{R}(s_0, \pi)$ of a policy $\pi$ in the state $s_0$ is the set of all states $s' \in S$ such that:
\begin{enumerate}
    \item $\pi(s') = \pi(s_0)$
    \item There exists a simple continuous path $\mathcal{P} = (s_0, \dots, s')$ such that for every state $s_j \in \mathcal{P}$, $\pi(s_j) = \pi(s_0)$, and each adjacent pair of states in the path is connected by an edge in $E_d$.
\end{enumerate}
\end{definition}
This definition ensures that $\mathcal{R}$ represents a locally connected "safe zone" of behavior, rather than disjoint pockets of identical actions.

\subsection{Characterizing the States within Robustness Regions}
\label{sec:characterizing-states-robustness-regions}

Importantly, the states in the Robustness Region range over the entire state space \(S\) of the Markov game, as represented by the user-defined factorization \(\mathcal{F}\). They are not restricted to states that would naturally arise from \(s_0\) along a standard RL trajectory.

An RL trajectory is generated by the agent’s policy \(\pi\) interacting with the environment, where each next state \(S_{t+1}\) is sampled from the transition function \(P(S_{t+1} \mid S_t, A_t)\). In contrast, a robustness region is built from hypothetical variations of \(s_0\) that may not be reachable from \(s_0\) by any finite sequence of transitions under \(P\). These variations are selected based on their "closeness" to \(s_0\), as measured by the hybrid distance \(d_{\text{hybrid}}\).

Because there is no inherent transition function linking an arbitrary perturbation \(s'\) directly to \(s_0\) in this abstract space, we require an alternative notion of connectivity. This motivates concepts such as \emph{immediate neighbors} to formally describe which states lie within the robustness region around \(s_0\).

\subsection{Minimal Counterfactual State}

\begin{definition}[Minimal Counterfactual State]
\label{def:min-counterfactual}
A minimal counterfactual state $s^*_{min}$ is a state such that $\pi(s^*_{min}) \neq \pi(s_0)$ that minimizes the distance to $s_0$:
\[
s^*_{min} \in \underset{s' \in S : \pi(s') \neq \pi(s_0)}{\operatorname{argmin}} \{ d_{\text{hybrid}}(s', s_0) \}
\]
\end{definition}

The set of all such states is denoted $\mathcal{C}_{\text{min}}(s_0, \pi)$. These states represent the "tipping points" where the policy is most sensitive to feature perturbations.

\begin{figure*}[t]
    \centering

    \begin{subfigure}[b]{0.47\textwidth}
        \includegraphics[width=\linewidth]{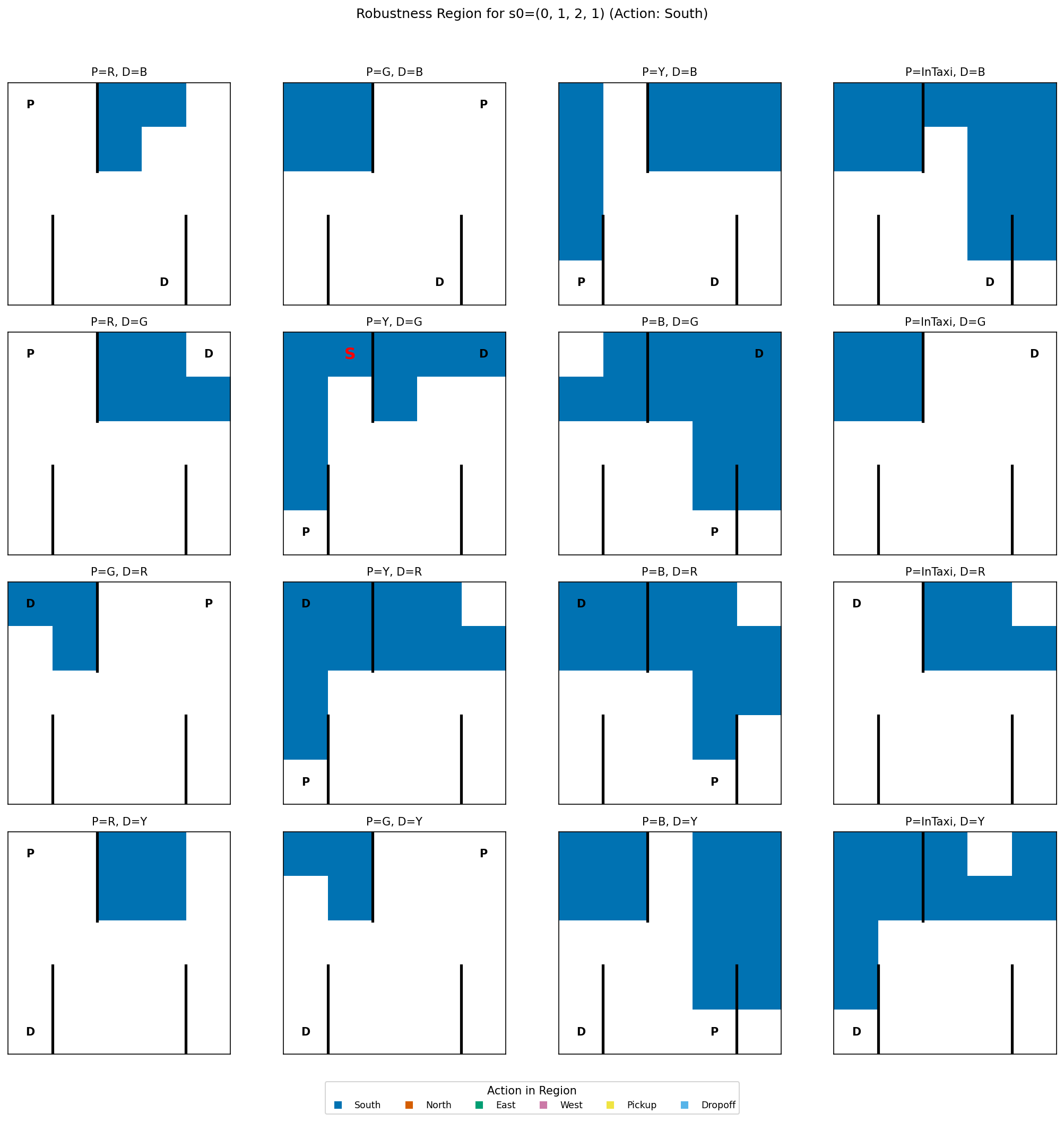}
        \caption{Robustness Region $\mathcal{R}(s_0, \pi)$. The uniform blue color across many subplots shows the action \textsc{South} is stable against changes in passenger (P) and destination (D) locations.}
        \label{fig:composite_rr_example}
    \end{subfigure}
    \hfill
    \begin{subfigure}[b]{0.47\textwidth}
        \includegraphics[width=\linewidth]{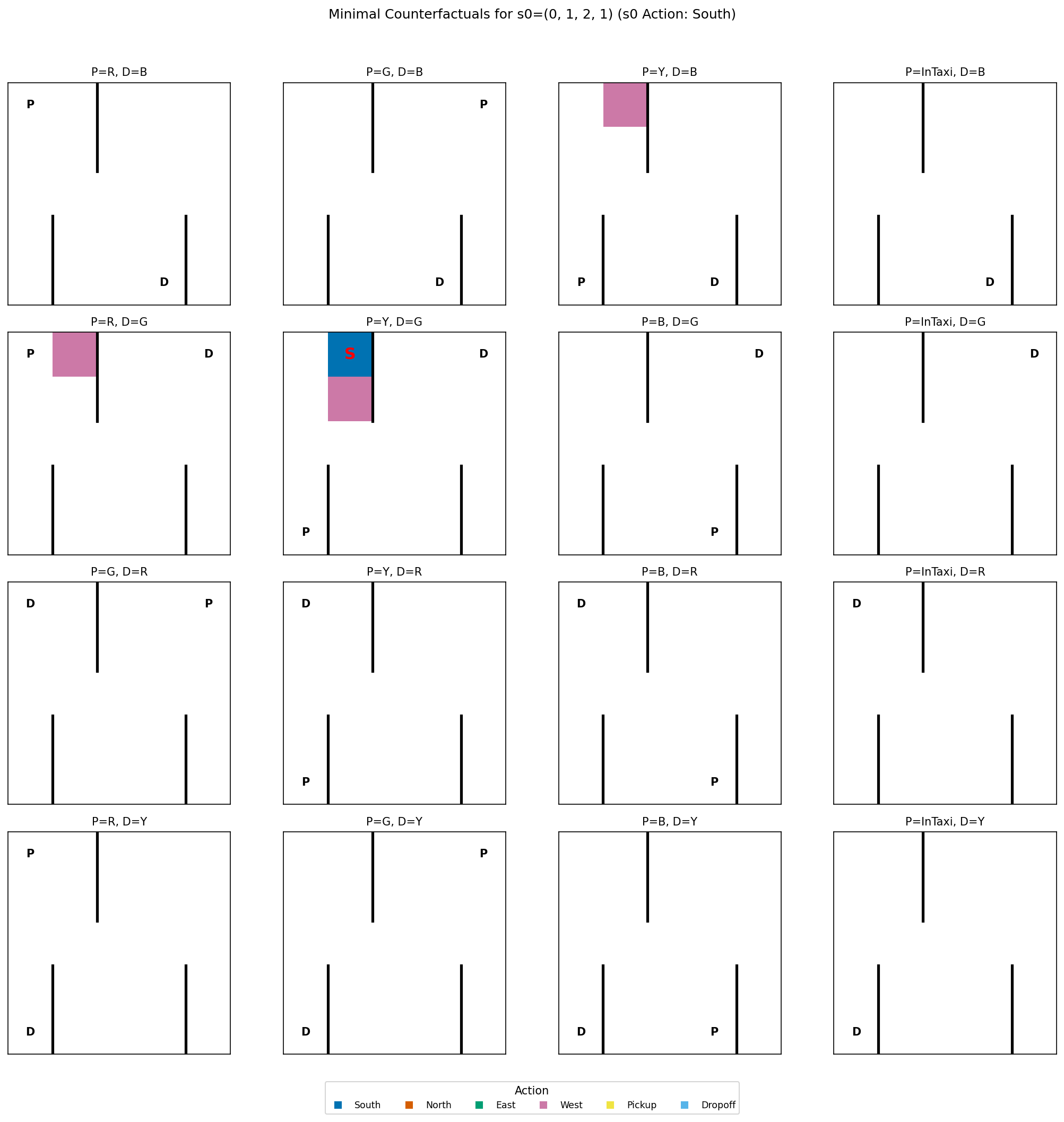}
        \caption{Minimal Counterfactuals $\mathcal{C}_{\text{min}}(s_0, \pi)$. In the relevant subplot (P=Y, D=G), the seed state ‘S’ is blue (\textsc{South}), while its neighbor is pink (\textsc{West}).}
        \label{fig:composite_cf_example}
    \end{subfigure}

    \caption{Visualisation of a composite explanation for an optimal policy
    $\pi_{\text{optimal}}$ in state $s_0=(0,1,2,1)$ (Taxi at (0,1), Passenger at Y,
    Dest at G). (a) The large robustness region shows the stability of the
    \textsc{South} action. (b) The minimal counterfactuals show the sensitive
    decision boundaries.}
    \label{fig:composite_explanation_example}
\end{figure*}

\subsection{Composite Explanations}

A Composite Explanation for a policy specific decision $\pi(s_0) = a_0$ consists of two parts:
\begin{enumerate}
    \item \textbf{Robustness Region ($\mathcal{R}$):} Addresses the question, "How stable is this decision?" by identifying the part of the induced graph around $s_0$ where the action remains invariant.
    \item \textbf{Minimal Counterfactuals ($\mathcal{C}_{\text{min}}$):} Addresses the question, "Why not another action?" by identifying the minimal perturbations required to flip the decision.
\end{enumerate}

Together, these components clarify both the extent of the policy's stability and the exact boundaries of its decision logic. An illustrative example of a composite explanation can be seen on Fig \ref{fig:composite_explanation_example}.

\subsection{Nature of Explanatory Robustness}
\label{sec:nature-robustness}

The states comprising a robustness region $\mathcal{R}(s_0, \pi)$ represent hypothetical variations of $s_0$ generated by perturbing state factors, rather than states necessarily reachable via the environment's transition dynamics $P(s' \mid s, a)$. While a standard RL trajectory follows probabilistic temporal transitions, our analysis traverses the \textit{state-similarity graph} induced by the factorization. Consequently, connectivity within a robustness region relies on metric closeness—sequences of $\delta_{\text{unit}}$ perturbations—rather than temporal reachability. This allows the framework to probe the policy's logic in counterfactual scenarios that may typically be inaccessible during standard episodic interaction.

\subsection{Algorithmic Computation}
We propose \textbf{STACHE-EXACT} (Algorithm \ref{alg:bfs}), a Breadth-First Search-based approach over the induced graph to compute these sets exactly. This approach leverages the layered exploration of BFS to guarantee both the connectedness of the robustness region and the minimality of the discovered counterfactuals.

\begin{algorithm}[tb]
   \caption{STACHE-EXACT}
   \label{alg:bfs}
\begin{algorithmic}
   \STATE {\bfseries Input:} State $s_0$, Policy $\pi$, Function $\texttt{GetNeighbors}$
   \STATE {\bfseries Output:} Region $\mathcal{R}$, Boundary Counterfactuals $C$
   \STATE Initialize $Q \gets [s_0]$, $V \gets \{s_0\}$
   \STATE $\mathcal{R} \gets \emptyset$, $C \gets \emptyset$
   \WHILE{$Q$ is not empty}
   \STATE $s' \gets Q.\text{pop}()$
   \IF{$\pi(s') = \pi(s_0)$}
       \STATE $\mathcal{R}.\text{add}(s')$
       \FOR{$s'' \in \texttt{GetNeighbors}(s')$}
           \IF{$s'' \notin V$}
               \STATE $V.\text{add}(s'')$
               \STATE $Q.\text{push}(s'')$
           \ENDIF
       \ENDFOR
   \ELSE
       \STATE $C.\text{add}(s')$
   \ENDIF
   \ENDWHILE
   \STATE \textbf{return} $\mathcal{R}, C$
\end{algorithmic}
\end{algorithm}

\textbf{Correctness.} BFS guarantees that states are visited in non-decreasing order of distance (number of unit steps). Thus, the algorithm finds the connected component of action-invariance ($\mathcal{R}$) and the set $C$ will contain all minimal counterfactuals reachable from $\mathcal{R}$. From $C$, we simply filter for those with minimal distance to $s_0$.

\textbf{Complexity.} The time complexity is $O(N \cdot T_{\pi} + M)$, where $N$ is the number of visited states (size of $\mathcal{R}$ + boundary), $M$ is the number of edges processed, and $T_{\pi}$ is the cost of a policy query. For massive state spaces, we employ a truncated version (\textbf{STACHE-CUTOFF}) that terminates once the first layer of minimal counterfactuals is fully resolved (details in Appendix \ref{app:stache-cutoff}).

\section{Experiments}
\label{sec:experiments}

We validate our framework on two discrete domains: \textbf{Taxi-v3} and \textbf{MiniGrid}. Through these experiments, we demonstrate how composite explanations can be used to diagnose policy maturity, quantify decision stability, and identify brittle logic in black-box agents.

\subsection{Taxi-v3: Evolution of Policy Logic}
\label{sec:experiments-taxi}

\textbf{Environment and Setup.} We utilize the canonical Taxi-v3 environment, modeled as a deterministic Markov Decision Process with a factored state space $S = (x, y, P, D)$. The state factors correspond to the taxi's coordinates $(x,y)$, the passenger location index $P$, and the destination index $D$. We analyze three Deep Q-Network (DQN) policies at distinct stages of training: $\pi_{0\%}$ (random/untrained), $\pi_{50\%}$ (intermediate), and $\pi_{100\%}$ (near-optimal). Our algorithm (STACHE-EXACT) uses the hybrid factored state distance to generate explanations.

\textbf{Analysis of Seed State $s_1 = (0,0,0,2)$.}
This state represents a critical decision point: the Taxi is at (0,0) (Landmark R), co-located with the passenger ($P=0$), while the destination is Y ($D=2$). The optimal action is \textsc{Pickup}. Table \ref{tab:taxi-s1} summarizes the quantitative evolution of the explanation.

\begin{table}[ht]
\caption{Evolution of explanations for Taxi state $s_1$ (Optimal Action: \textsc{Pickup}). Comparison of Action choice, Robustness Region (RR) size, and the nature of Minimal Counterfactuals (CF).}
\label{tab:taxi-s1}
\vskip 0.15in
\begin{center}
\begin{small}
\begin{sc}
\begin{tabular}{lccc}
\toprule
Metric & $\pi_{0\%}$ & $\pi_{50\%}$ & $\pi_{100\%}$ \\
\midrule
Action & \textsc{North} (Invalid) & \textsc{Pickup} & \textsc{Pickup} \\
RR Size & 9 states & 3 states & 3 states \\
CF Logic & Random/Chaotic & Specific & Specific \\
\bottomrule
\end{tabular}
\end{sc}
\end{small}
\end{center}
\vskip -0.1in
\end{table}

\begin{figure*}[t]
    \centering
    \begin{subfigure}[b]{0.31\textwidth}
        \includegraphics[width=\linewidth]{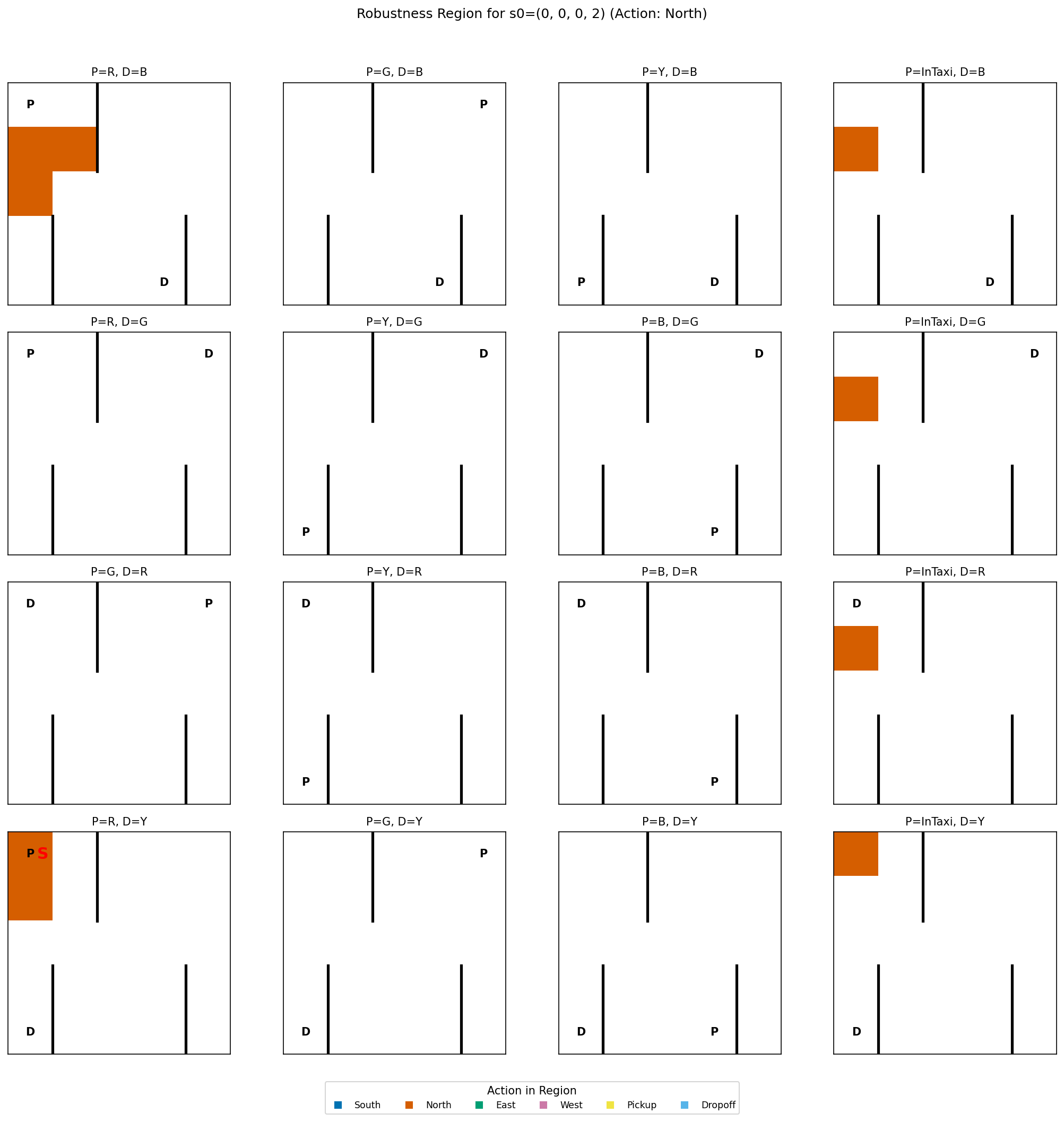}
        \caption{$\pi_{0\%}$: RR Size 9. The policy selects \textsc{North} (Orange).}
        \label{fig:taxi_s1_0}
    \end{subfigure}
    \hfill
    \begin{subfigure}[b]{0.31\textwidth}
        \includegraphics[width=\linewidth]{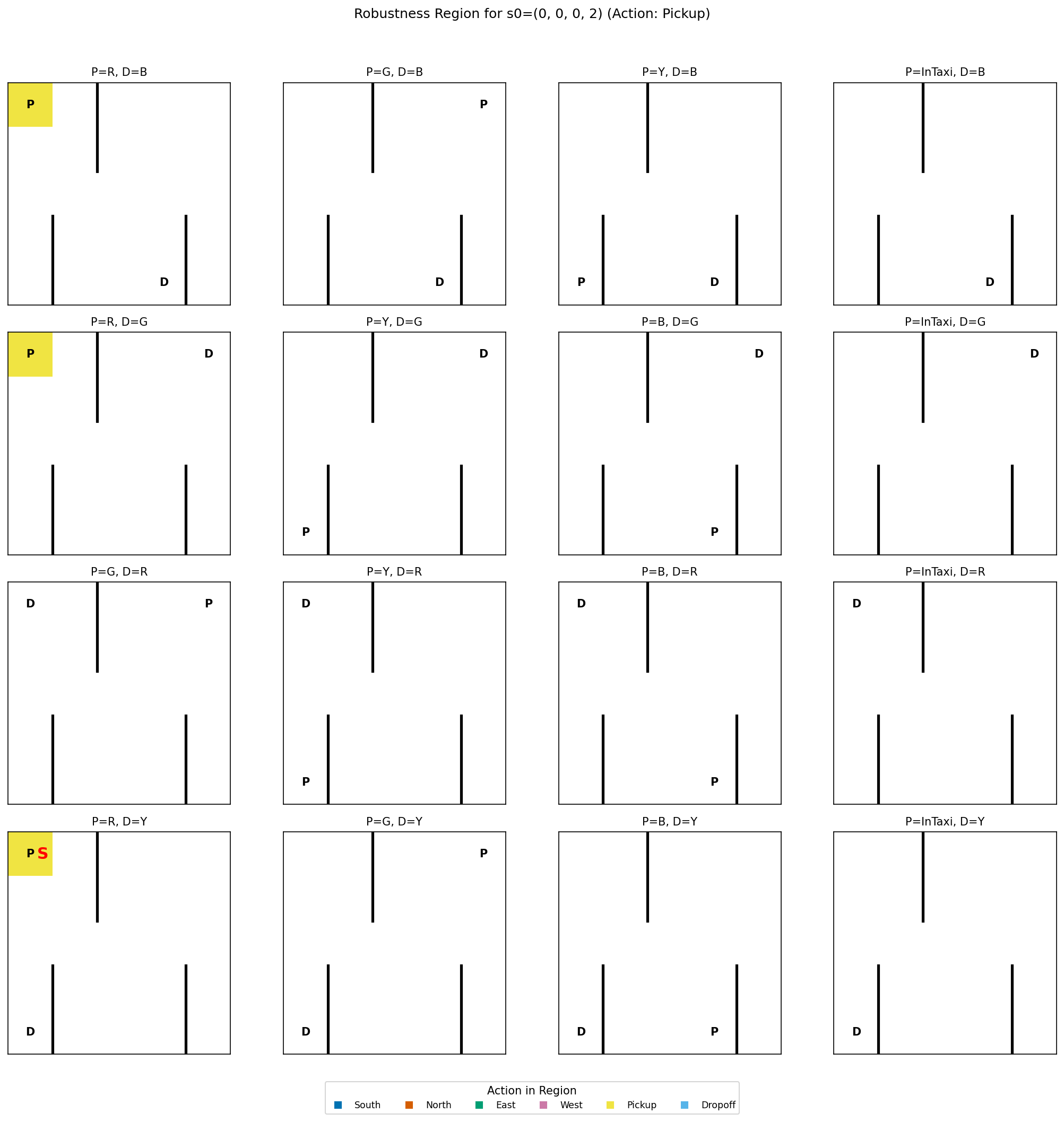}
        \caption{$\pi_{50\%}$: RR Size 3. The policy correctly selects \textsc{Pickup} (Brown).}
        \label{fig:taxi_s1_50}
    \end{subfigure}
    \hfill
    \begin{subfigure}[b]{0.31\textwidth}
        \includegraphics[width=\linewidth]{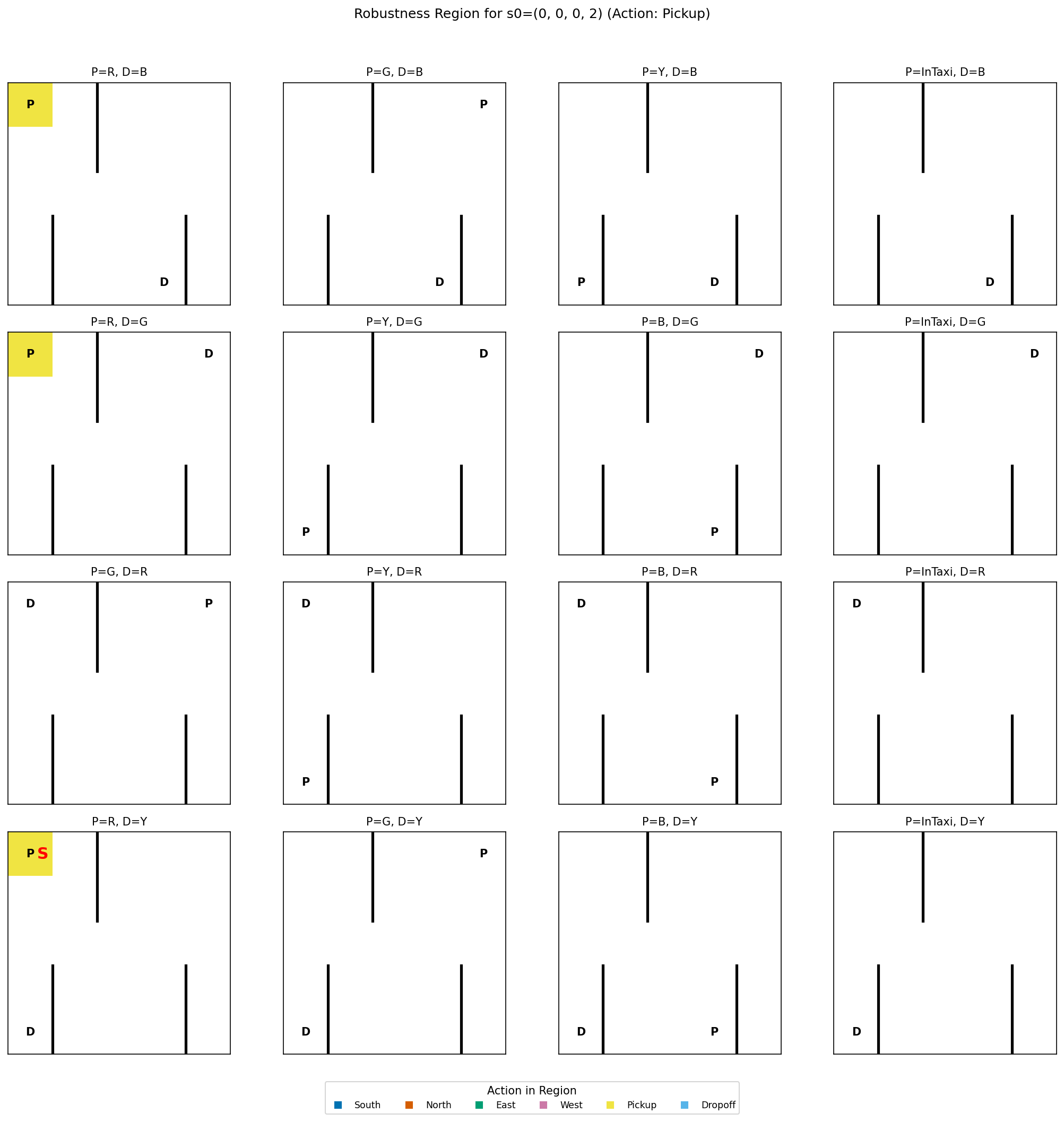}
        \caption{$\pi_{100\%}$: RR Size 3. Action remains \textsc{Pickup} (Brown).}
        \label{fig:taxi_s1_100}
    \end{subfigure}
    \caption{Evolution of Robustness Regions for state $s_1=(0,0,0,2)$ across training stages. Colored cells indicate stability (same action as seed). As the agent learns the highly specific \textsc{Pickup} logic, the robustness region shrinks from a random collection of states (a) to a precise configuration (b, c), demonstrating that for critical actions, high performance correlates with low robustness region size (high specificity).}
    \label{fig:taxi_s1_evolution}
\end{figure*}

As shown in Figure \ref{fig:taxi_s1_evolution}, the nature of the explanation shifts dramatically:
\begin{itemize}
    \item \textbf{$\pi_{0\%}$ (Untrained):} The agent selects \textsc{North}, an invalid action driving into a wall. The RR size is 9, including intuitively unrelated states. Minimal CFs are chaotic; single-factor perturbations lead to arbitrary actions like \textsc{West} or \textsc{East}.
    \item \textbf{$\pi_{50\%}$ \& $\pi_{100\%}$ (Trained):} Both policies converge on \textsc{Pickup}. Crucially, the RR shrinks to just 3 states. This small region contains only variations in the \textit{destination} factor ($D$), which is irrelevant to the immediate legality of a pickup. This indicates \textbf{high specificity}: the explanation reveals that the agent has learned that \textsc{Pickup} is brittle and valid only under strict spatial conditions. The minimal CFs become logically coherent: shifting the taxi ($x: 0 \to 1$) or the passenger ($P: 0 \to 1$) immediately flips the action to a navigational move (\textsc{North} or \textsc{South}), confirming the policy's sensitivity to task-critical features.
\end{itemize}

\textbf{Analysis of Navigation State $s_2 = (0,1,2,1)$.}
To contrast with the specificity of pickup actions, we analyze a navigational state where the Taxi is at (0,1) and must travel to the Passenger at Y ($P=2$).
\begin{itemize}
    \item \textbf{$\pi_{50\%}$ (Partial):} The agent incorrectly selects \textsc{North} (moving away from the target). The RR size is 1. This extreme instability indicates a "confused" policy region where any perturbation alters the decision.
    \item \textbf{$\pi_{100\%}$ (Optimal):} The agent correctly selects \textsc{South}. The RR size expands to 125 states. Unlike the pickup case, here a \textit{larger} RR indicates mastery. The explanation shows the agent maintains the \textsc{South} bearing across a vast range of passenger/destination configurations, provided the taxi remains in the upper grid zone.
\end{itemize}

\textbf{Key Finding:} We observe distinct evolutionary patterns depending on the action type. As training progresses, Robustness Regions for specific actions (Pickup/Dropoff) \textit{shrink} (reflecting necessary precision), while RRs for navigational actions \textit{grow} (reflecting generalization and stability).

\subsection{MiniGrid: Diagnosing Policy Brittleness}
\label{sec:experiments-minigrid}

\textbf{Environment.} We apply our framework to \texttt{MiniGrid-Empty-Random-6x6}, a sparse-reward navigation task. The state is factored into the Agent's pose (position and direction) and the Goal position. We analyze a PPO agent to identify regions of stability versus brittleness.

\textbf{Corridors of Stability ($s_{36}$).}
In states where the agent is aligned with the goal (e.g., Agent at (1,2) facing Down, Goal at (4,4)), the PPO agent selects \textsc{Move Forward}. The computed RR size is 16. Visualizing the RR reveals a "corridor" of valid states: the agent persists in moving forward even if laterally displaced, provided the general heading remains viable. Minimal CFs in this scenario correspond to 1-step lateral shifts that logically force a \textsc{Turn} action to realign.

\begin{figure}[t]
    \centering
    \begin{subfigure}[b]{0.3\columnwidth}
        \includegraphics[width=\linewidth]{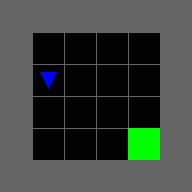}
        \caption{}
        \label{fig:minigrid_brittle_rr_down}
    \end{subfigure}
    \hfill
    \begin{subfigure}[b]{0.3\columnwidth}
        \includegraphics[width=\linewidth]{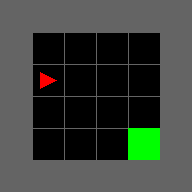}
        \caption{}
        \label{fig:s38_rr_dir0}
    \end{subfigure}
    \hfill
    \begin{subfigure}[b]{0.3\columnwidth}
        \includegraphics[width=\linewidth]{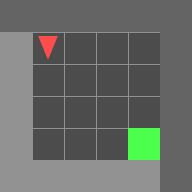}
        \caption{}
        \label{fig:minigrid_brittle_cf}
    \end{subfigure}
    \caption{Diagnosing policy brittleness in MiniGrid state $s_{38}$. 
    (a) Robustness region map when the agent faces \textsc{Down}. Blue arrow indicates initial position.
    (b) Robustness region map for the same state when the agent faces \textsc{Right}. 
    Robustness region is extremely small, indicating that the action 
    \textsc{Move Forward} is highly unstable at this location. 
    (c) A minimal counterfactual illustrating that simply moving the agent one cell up causes the action to flip to \textsc{Right}, revealing a brittle change in strategy.}
    \label{fig:minigrid_brittle}
\end{figure}

\textbf{Identifying Brittleness ($s_{38}$).}
Conversely, we identified states such as $s_{38}$ (Agent at (2,1), Goal (4,4)) where the policy seemingly chooses \textsc{Move Forward}, but the explanation reveals deep fragility.
As shown in Figure \ref{fig:minigrid_brittle}, the RR size drops to 2. This signifies that almost any perturbation—changing the direction or shifting position by a single cell—causes the agent to abandon the \textsc{Forward} action.
The minimal counterfactuals (Figure \ref{fig:minigrid_brittle_cf}) highlight that the agent is on a "razor's edge": minor environmental noise would cause it to switch to \textsc{Turn Left} or \textsc{Turn Right}. This characterizes a lack of confidence in the policy for that specific sector of the state space, a diagnostic insight unavailable through reward curves alone.

\section{Discussion and Limitations}
\label{sec:discussion}

\textbf{Utility.} The framework successfully distinguishes between "robust" decisions (large RR) and "brittle" ones (small RR). It provides debugging value by exposing exactly \textit{which} factors break a decision (Minimal CFs). For example, in MiniGrid, we identified cases where the agent would turn into a wall if its orientation changed, a flaw hidden by standard evaluation.

\textbf{Limitations.}
1. \textit{Scalability:} The exact BFS scales with the size of the robustness region. For large regions in complex games, the truncated BFS (Appendix \ref{app:stache-cutoff}) is necessary.
2. \textit{Feature Granularity:} The explanations are only as good as the user-defined factorization. If factors are not semantic (e.g., raw pixels), the resulting CFs may be less interpretable.

\section{Conclusion}
\label{sec:conclusion}

We presented a method for explaining discrete RL policies using Robustness Regions and Minimal Counterfactuals. By treating the policy as a black box and exploring the factored state space, we provide exact, verifiable insights into decision stability and sensitivity. Our experiments show that these metrics track the "crystallization" of agent logic during training. Our future work will focus on scaling these exact guarantees. Key directions include: (1) extending local regions to temporal "robustness tubes" for trajectory explanations; (2) deriving local safety certificates from region boundaries; and (3) updating the algorithm to leverage modern SMT solvers to accelerate computation without compromising the exactness of the results.


\bibliography{arxiv_bibliography}
\bibliographystyle{plainnat}

\newpage
\appendix
\onecolumn
\section{Additional Experimental Details}

\subsection{Environment Details}

\textbf{Taxi-v3.} The state space consists of 500 discrete states (404 reachable).
State factors:
\begin{itemize}
    \item Taxi Row $x \in \{0..4\}$ (Numerical)
    \item Taxi Col $y \in \{0..4\}$ (Numerical)
    \item Passenger Index $P \in \{0..4\}$ (Categorical: 0=R, 1=G, 2=Y, 3=B, 4=In Taxi)
    \item Destination Index $D \in \{0..3\}$ (Categorical)
\end{itemize}
Distance Metric: $d((x,y,P,D), (x',y',P',D')) = |x-x'| + |y-y'| + \mathbb{I}(P \neq P') + \mathbb{I}(D \neq D')$.

\textbf{MiniGrid.} We used `MiniGrid-Empty-Random-6x6-v0`.
State factors used for explanation:
\begin{itemize}
    \item Agent Pos $(x,y)$ (Numerical)
    \item Agent Dir (Categorical/Cyclic - treated as categorical for simple distance)
    \item Goal Pos $(x,y)$ (Numerical)
\end{itemize}

\section{Extended Algorithmic Details}
\label{app:algorithms}

\subsection{Truncated Search for Scalability (STACHE-CUTOFF)}
\label{app:stache-cutoff}

For environments with large state spaces, exhaustively mapping the full robustness region $\mathcal{R}(s_0, \pi)$ may be computationally prohibitive. However, finding the \textit{minimal} counterfactuals often requires exploring only a small radius around $s_0$. We employ a cutoff-based version of STACHE (Algorithm \ref{alg:cutoff}) that prioritizes discovering the closest reasons for policy change.

\begin{algorithm}[ht]
   \caption{STACHE-CUTOFF}
   \label{alg:cutoff}
\begin{algorithmic}[1]
   \STATE {\bfseries Input:} State $s_0$, Policy $\pi$, Function $\texttt{GetNeighbors}$
   \STATE {\bfseries Output:} Partial Region $\mathcal{R}'$, Minimal Counterfactuals $\mathcal{C}_{\min}$
   \STATE Initialize $Q \gets [(s_0, 0)]$, $V \gets \{s_0\}$
   \STATE $\mathcal{R}' \gets \emptyset$, $\mathcal{C}_{\min} \gets \emptyset$
   \STATE $\text{min\_dist} \gets \infty$
   \WHILE{$Q$ is not empty}
       \STATE $(s', d') \gets Q.\text{pop}()$
       \IF{$d' > \text{min\_dist}$}
           \STATE \textbf{continue}
       \ENDIF
       \IF{$\pi(s') \neq \pi(s_0)$}
           \IF{$d' < \text{min\_dist}$}
               \STATE $\text{min\_dist} \gets d'$
               \STATE $\mathcal{C}_{\min} \gets \{s'\}$
           \ELSIF{$d' = \text{min\_dist}$}
               \STATE $\mathcal{C}_{\min}.\text{add}(s')$
           \ENDIF
       \ELSE
           \STATE $\mathcal{R}'.\text{add}(s')$
           \FOR{$s'' \in \texttt{GetNeighbors}(s')$}
               \STATE $d_{\text{next}} \gets d' + 1$
               \IF{$d_{\text{next}} \le \text{min\_dist}$ \AND $s'' \notin V$}
                   \STATE $V.\text{add}(s'')$
                   \STATE $Q.\text{push}((s'', d_{\text{next}}))$
               \ENDIF
           \ENDFOR
       \ENDIF
   \ENDWHILE
   \STATE \textbf{return} $\mathcal{R}', \mathcal{C}_{\min}$
\end{algorithmic}
\end{algorithm}

This algorithm ensures that we find \textit{all} counterfactuals at the minimal distance $\rho_d(s_0, \pi)$ but prevents unnecessary expansion into the deep interior of the robustness region beyond this distance.

\subsection{Complexity Analysis}
Let $N_{\mathcal{R}} = |\mathcal{R}(s_0, \pi)|$ be the number of states in the robustness region, and $N_C$ be the number of boundary counterfactuals.
\begin{itemize}
    \item \textbf{Time Complexity:} The algorithm visits each state in the region and its immediate boundary once. For each visited state, it performs one policy query $T_{\pi}$ and one neighbor generation step $T_{\text{GetN}}$. The total time is $O((N_{\mathcal{R}} + N_C) \cdot (T_{\pi} + T_{\text{GetN}}))$.
    \item \textbf{Space Complexity:} We must store the visited set $V$ and the queue $Q$. The space requirement is $O(N_{\mathcal{R}} + N_C)$.
\end{itemize}
For the \textbf{Cutoff} variant, let $\rho^*$ be the minimal counterfactual distance. The complexity is reduced to the number of states $N'$ within distance $\rho^*$ of $s_0$, i.e., $O(N' \cdot T_{\pi})$.

\section{Theoretical Guarantees}
\label{app:theory}

We provide formal justification for the correctness of the proposed methodology using the definitions provided in Section \ref{sec:methodology}.

\textbf{Proposition 1 (Robustness Region Correctness).}
\textit{Any state $s$ added to the set $\mathcal{R}$ by Algorithm \ref{alg:bfs} satisfies the definition of a Robustness Region (Definition \ref{def:robustness-region}).}

\textit{Proof.}
The algorithm adds $s$ to $\mathcal{R}$ only if $\pi(s) = \pi(s_0)$. Furthermore, because the search strategy is BFS starting from $s_0$, any state $s$ added to the queue implies the existence of a parent state $s_{parent}$ which was already in $\mathcal{R}$. By induction, there exists a chain of immediate neighbors $(s_0, s_1, \dots, s)$ where each $s_i$ is connected by a distance of $\delta_{\text{unit}}$ and satisfies the policy invariance condition. This forms the \textit{simple continuous path} required by Definition \ref{def:robustness-region}.

\textbf{Proposition 2 (Minimal Counterfactual Identification).}
\textit{The set $\mathcal{C}_{\min}$ returned by Algorithm \ref{alg:cutoff} contains all minimal counterfactual states and no non-minimal counterfactuals.}

\textit{Proof.}
BFS explores states in strictly non-decreasing order of distance $d(s_0, \cdot)$.
Let $\rho^*$ be the true minimal distance to any state $s'$ with $\pi(s') \neq \pi(s_0)$.
\begin{enumerate}
    \item The algorithm will eventually deque states at distance $\rho^*$. Since $\rho^*$ is minimal, no counterfactuals would have been found at distances $d < \rho^*$, so `min\_dist` remains $\infty$.
    \item Upon encountering the first counterfactual $s^*$ at $d(s^*)=\rho^*$, `min\_dist` is updated to $\rho^*$.
    \item The algorithm continues to process all other states at distance $\rho^*$ in the current queue layer, adding any valid counterfactuals to $\mathcal{C}_{\min}$.
    \item Any states at distance $d > \rho^*$ are pruned by the condition `if $d' > \text{min\_dist}$`, ensuring no non-minimal counterfactuals are added.
\end{enumerate}
Thus, $\mathcal{C}_{\min}$ exhaustively contains the "tipping points" of the policy.

\end{document}